\documentclass{article}

\usepackage[utf8]{inputenc}
\usepackage{colm2024_conference}

\usepackage{graphicx}
\usepackage{enumitem}
\usepackage{wrapfig}
\usepackage{algorithm}
\usepackage{algpseudocode}
\usepackage{natbib}
\usepackage{makecell}
\usepackage{amsmath}
\usepackage{amssymb}
\usepackage{amsfonts}
\usepackage{multirow}
\usepackage{verbatim}
\usepackage{caption}
\usepackage{longtable}
\usepackage{supertabular}
\usepackage{float}
\usepackage[section]{placeins}
\usepackage{pifont}

\usepackage{enumitem}
\usepackage{tablefootnote}
\usepackage{xspace}
\usepackage{textcomp}
\usepackage{makecell}
\usepackage{lscape}
\usepackage{siunitx}
\usepackage{listings}
\usepackage{xcolor}

\definecolor{dt}{gray}{0.7}
\usepackage{pifont}
\usepackage{bbding}
\usepackage{fontawesome}

\usepackage{scrextend}
\usepackage{ulem}
\usepackage{tgpagella}
\usepackage{latexsym}
\usepackage[T1]{fontenc}
\usepackage{microtype}
\definecolor{mydarkblue}{rgb}{0,0.08,0.45}
\definecolor{citecolor}{HTML}{0071BC}
\usepackage{url}
\usepackage{nicefrac}
\usepackage{changepage}
\usepackage{xargs}
\usepackage{wrapfig,lipsum,booktabs}
\usepackage{longtable}
\usepackage{subcaption}
\usepackage{endnotes}

\usepackage{pgfplots}
\usetikzlibrary{pgfplots.groupplots}
\pgfplotsset{compat=1.3}
\usepackage{tikz}
\usetikzlibrary{patterns}

\usepackage[most]{tcolorbox}

\usepackage{hyperref}
\definecolor{darkblue}{rgb}{0, 0, 0.5}
\hypersetup{colorlinks=true, citecolor=darkblue, linkcolor=red, urlcolor=darkblue, bookmarksopen=false, bookmarksnumbered=true}

\usepackage[capitalize,noabbrev]{cleveref}

\crefname{section}{\S}{\S\S}
\Crefname{section}{\S}{\S\S}
\crefname{subsection}{\S\S}{\S\S}
\Crefname{subsection}{\S\S}{\S\S}
\crefformat{subsection}{\S\S#2#1#3}

\crefname{table}{Table}{Tables}
\crefname{figure}{Figure}{Figures}
\crefname{algorithm}{Algorithm}{}
\crefname{equation}{eq.}{}
\crefname{appendix}{Appendix}{}
\crefformat{section}{Section #2#1#3}
\usepackage{multicol}
\usepackage{tcolorbox}
\usepackage{titlesec}
\titleformat*{\section}{\large\bfseries}
\newcolumntype{P}[1]{>{\centering\arraybackslash}p{#1}}
\usepackage{multirow}

\usepackage{adjustbox}
\definecolor{objblue}{RGB}{3,139,221}
\definecolor{attrred}{RGB}{255,67,67}
\definecolor{easygreen}{RGB}{0,156,75}
\definecolor{middleyellow}{RGB}{242,89,34}
\definecolor{hardred}{RGB}{216,56,58}
\definecolor{morandigreen}{RGB}{91,122,104}
\usepackage{colortbl}
\usepackage{array}

\usepackage[most]{tcolorbox}
\definecolor{BoxBackground}{RGB}{240, 240, 240}
\definecolor{BoxFrame}{RGB}{0, 0, 0}
\definecolor{TitleBackground}{RGB}{0, 0, 0}
\definecolor{TitleText}{RGB}{255, 255, 255}
\tcbset{
  academicbox/.style={
    boxsep=5pt,
    left=2pt,
    right=2pt,
    bottom=0.5pt,
    boxrule=0.5pt,
    colback=BoxBackground,
    colframe=BoxFrame,
    colbacktitle=TitleBackground,
    coltitle=TitleText,
    enhanced,
    attach boxed title to top left={yshift=-0.1in,xshift=0.1in},
    boxed title style={boxrule=0pt,colframe=white},
    title={#1},
  }
}
\newtcolorbox{AcademicBox}[1][]{academicbox=#1}

\graphicspath{{image/}}

\title{CommerceGen-Studio: An End-to-End Framework for Commercial-Grade Main-Image Generation}
\author{
{\small \textit{AI-Native Applications Group}} \\[1mm]
Xiaoli Xu,
Zhengtao Wu,
Zhengze Xu,
Tao Liu,
Yongchao Du,
Zuan Gao,
Xingjian Wang,
Hao Yan,
Mengting Chen\textsuperscript{*},
Wei Li,
Bo Zheng \\[2mm]
\textbf{Alibaba Group} \\[2mm]
{\small
\textsuperscript{*}Project Lead 
}
}

\begin{document}

\maketitle

\begin{abstract}
We present CommerceGen-Studio, an end-to-end framework for generating commercial-grade main images for e-commerce Stock Keeping Units (SKUs). Our framework is organized along a
three-stage pipeline of information extraction, semantic
rewriting, and multi-condition generation: Prompt-Eval-Refine-Loop Agent (PERL-Agent) distills heterogeneous SKU metadata into concise selling-point text lists through an iterative closed loop of VLM and LLM roles scored by a reference-free rubric. Layout Thinker Optimization (LTO) aligns the rewriter with both caption and rendered-image quality via alternating domain-specific policy optimization. SKU-DiT, trained under Reinforce-Distill-Reinforce (RDR) progressive training, conditions jointly on the white-background product image, the trademark crop, and the structured layout to hold subject fidelity, trademark integrity, and character-level text accuracy together. The framework runs in production at a scale of millions of main images per day with a latency of a few seconds per image, which we believe makes it among the first systems of this kind to operate at commercial scale. We also release CommerceGen-Bench, a benchmark of hard cases collected from live traffic across all product categories. On it, our framework achieves the highest overall scores among the evaluated open-source baselines on both editing tracks, with and without prompt enhancement. It also matches much larger closed-source models, and beats several of them at  Instruction Following and  Product Integration - the two properties that determine whether an image can be published.
\end{abstract}

\section{Introduction}

On e-commerce search result pages (SRPs), shoppers judge relevance almost entirely from two signals: the product image and the title. As search engines increasingly resolve a query to one specific Stock Keeping Unit (SKU) rather than to an item containing multiple SKUs, the SRP has to render display assets at the precise SKU granularity rather than the coarser item granularity. Merchants, however, rarely maintain them. SKU-level imagery is dominated by plain white-background product photographs, and dropping such photographs directly into the SRP measurably lowers click-through rate. Therefore, automated production of high-quality main images at SKU granularity is both a critical need for experience optimization and an inevitable choice for supply gap remediation.

General-purpose text-to-image and image-to-image foundation models have advanced rapidly \cite{hu2026orion,hu2026swiftimage,chen2024referenceimitation}; \cite{wu2025qwenimage,ma2025longcatimage,song2026joyai,qiao2026firered,blackforestlabs2025flux2,liu2026postertext}, yet deploying them for SKU product cards exposes four requirements they were not designed to meet. First, selling points must be constructed from heterogeneous sources. SKU information is scattered across titles, OCR results from product images, structured attributes, and backend parameters. These sources may overlap, conflict, or be incomplete, and the relevant information varies by category. They must therefore be reconciled into a concise and reliable set of selling points before generation begins. Second, information consistency is essential: subjects, trademarks, and selling-point text must match the source SKU metadata exactly, since even a hallucinated digit or conflicting attribute may mislead a purchase decision. Third, long-tail robustness is required because live traffic spans diverse product categories and merchant images of highly uneven quality. Fourth, multi-condition layout control must jointly coordinate the product, trademark, and selling-point text within a coherent composition. Existing methods address these requirements only partially: cascaded ``plan-then-render'' systems accumulate errors across stages, while end-to-end poster models often trade text accuracy against 

\begin{figure}[H]
    \centering
    \includegraphics[width=0.95\linewidth]{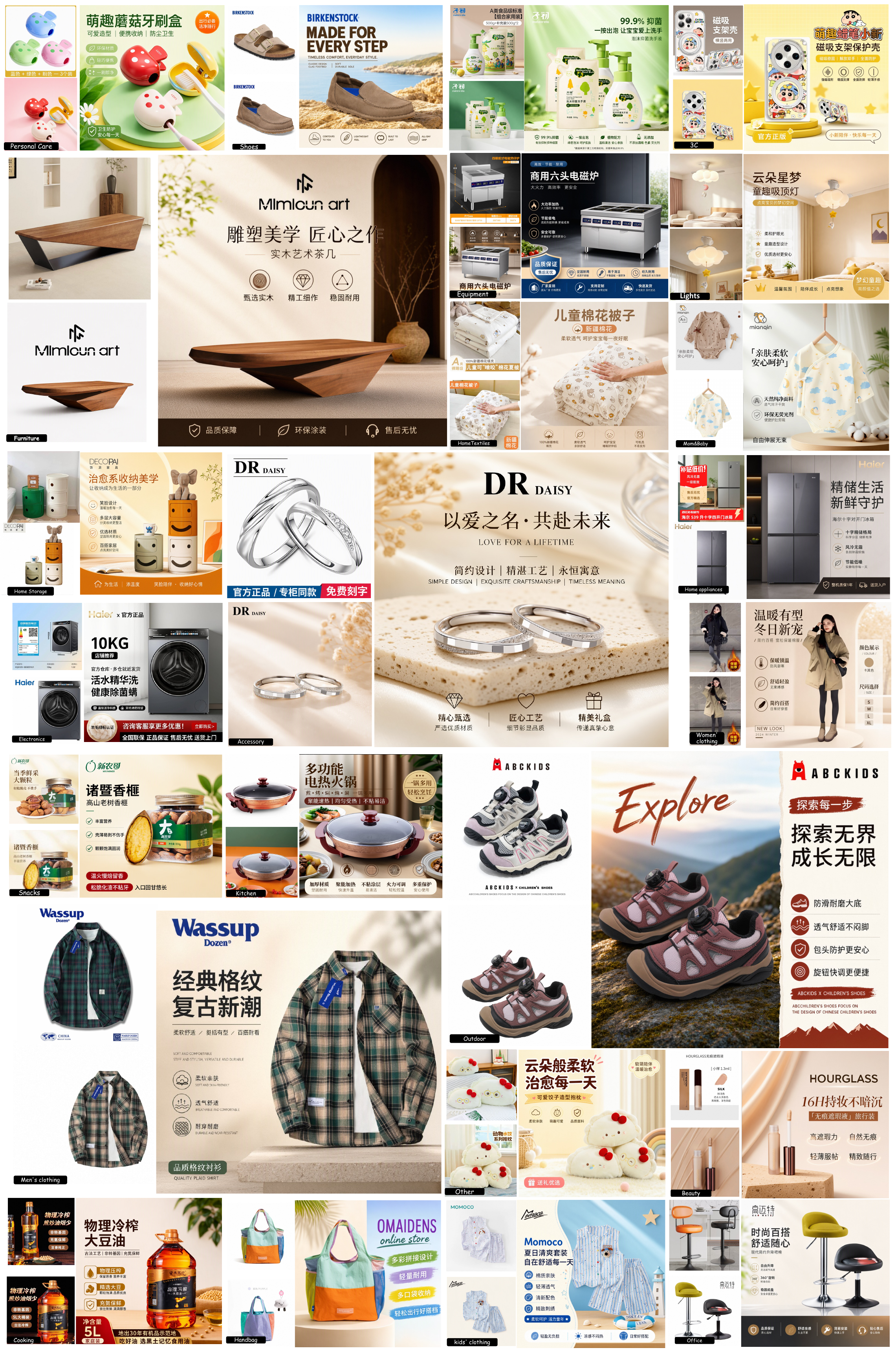}
    \caption{\textbf{Showcase of CommerceGen-Studio}. Given a product main image, an SKU image, and structured product information, the system automatically generates commercial-grade images with subject fidelity, trademark consistency, and precise selling-point text rendering.}
    \label{fig:showcase}
\end{figure}

subject fidelity. Figure~\ref{fig:showcase} illustrates our target setting: given a primary product image and an original SKU image, the system must produce a ready-to-use SKU image that satisfies all four requirements.

We address these requirements with a system designed end to end rather than assembled from independently tuned parts. Our contributions are as follows. (1) We present CommerceGen-Studio, a fully automated production framework organized around three stages -- information extraction, semantic rewriting, and multi-condition generation -- that turns heterogeneous SKU metadata into executable 
layout semantics and renders it with a single multi-image editing Diffusion Transformer (DiT). (2) We present Prompt-Eval-Refine-Loop Agent (PERL-Agent), which recasts selling-point extraction as an iterative closed loop over four roles (Generator, Judge, Analyzer, and Refiner; one VLM and three LLMs) scored by a six-dimensional reference-free rubric. One category-adaptive prompt serves every category, and attribution over the dominant failure dimension lets the loop roll back to a specific weakness instead of restarting. 
(3) We present Layout Thinker Optimization (LTO), a VLM layout rewriter trained on two independently sampled reward domains -- one scoring captions directly and one scoring rendered images. Rewards are normalized within each domain's rollout groups, the two policy losses are computed separately, and each optimization step samples one reward domain for backpropagation according to the configured domain probability. (4) We present SKU-DiT together with Reinforce-Distill-Reinforce (RDR) progressive training, which sequences preference alignment, Dynamic Backward Simulation Distillation, and a second round of reinforcement on the few-step student, reaching a few seconds latency without conceding the text precision and subject fidelity earned earlier. (5) We release CommerceGen-Bench, a hard-sample benchmark spanning all e-commerce categories, with a five-dimensional rubric scored by a multimodal judge model so that results on it are reproducible across systems.


\section{Related Work}
\subsection{E-commerce Poster Generation}

\begin{figure}[!t]
\centering
\includegraphics[width=\linewidth]{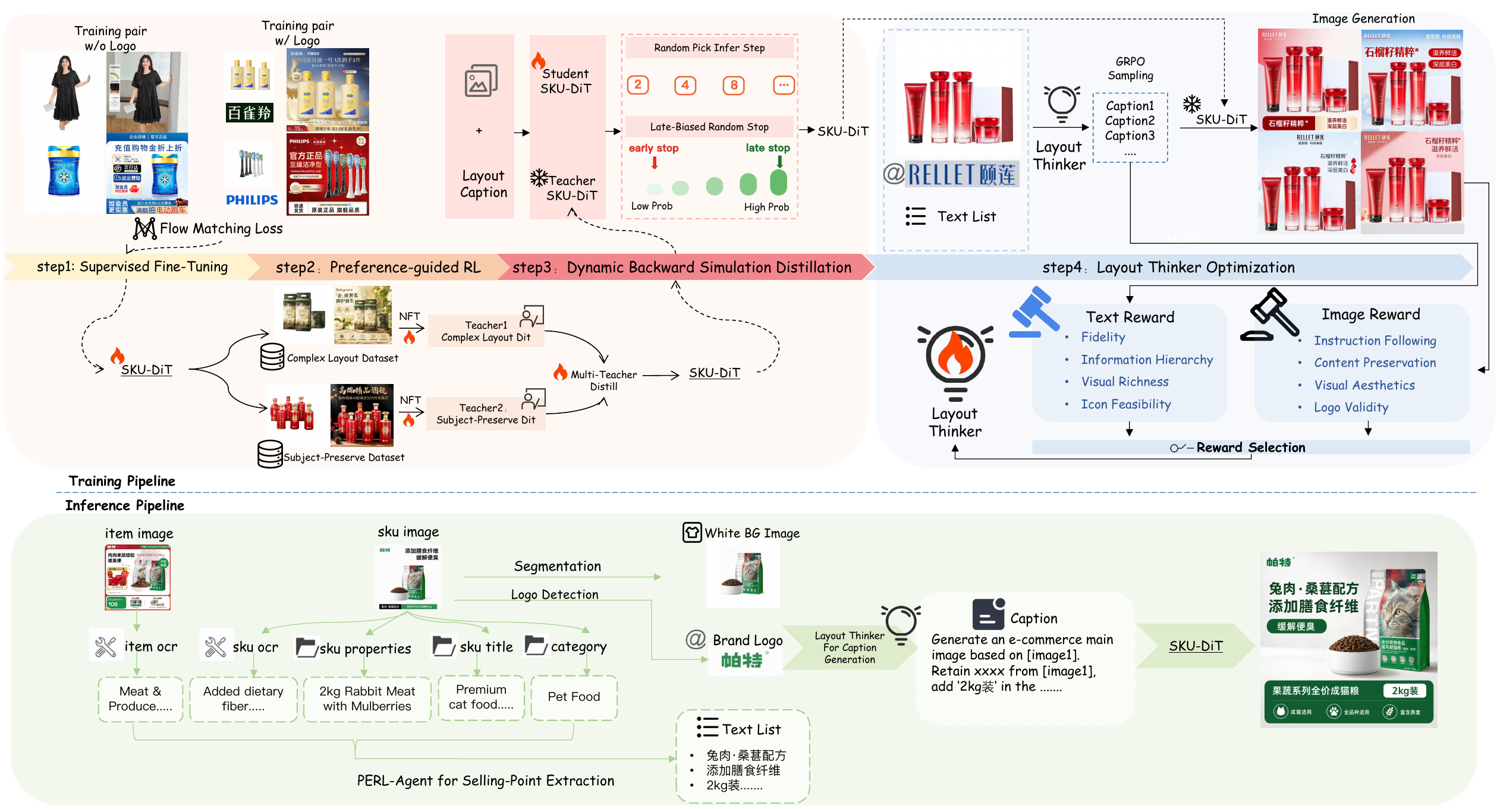}
\caption{Overall framework of CommerceGen-Studio. The training pipeline combines supervised fine-tuning, RDR progressive training, and LTO. The inference pipeline performs selling-point text list  extraction, Layout Thinker rewriting, and multi-condition SKU-DiT generation. Dashed lines connect training-time components to their inference-time roles.}
\label{fig:pipeline}
\end{figure}

Recent e-commerce poster systems have moved beyond fixed templates toward learned composition and text rendering. PosterMaker \cite{gao2025postermaker} and PosterCraft \cite{chen2025postercraft} generate complete posters with accurate or stylized typography, but PosterCraft starts from a text prompt rather than a product reference, while PosterMaker does not address heterogeneous SKU metadata or separate trademark conditioning. Among 2026 systems, InnoAds-Composer \cite{qin2026innoads} jointly conditions on a subject image, a glyph image, and a style reference, yet receives the text and approximate placement as conditions instead of extracting selling points or planning a complex layout. AutoPP \cite{fan2026autopp} predicts background, selected copy, and layout from one product image and a supplied candidate-copy set, then optimizes the design model with online click feedback; it selects from prepared copy rather than reconciling raw titles, OCR, attributes, and parameters, and it has no separate trademark reference. PosterVerse \cite{liu2026posterverse} converts a user request into a blueprint and editable HTML poster with scalable typography, but its interface is prompt-driven and does not preserve multiple SKU reference assets. PosterText \cite{liu2026postertext} unifies poster creation and editing by treating text patches as atomic units; it accepts one primary product image or poster together with an optional set of reference patches, but does not extract selling points from heterogeneous SKU sources or provide a metadata-driven adaptive layout planner. TransAnyText \cite{liu2026transanytext} localizes an existing e-commerce image by translating its text and generating an HTML patch whose position, font, color, and size adapt to the target language. Its documented inference input is a single source image, and its objective is localization rather than SKU-level selling-point construction or multi-reference product generation.

Reference-guided editing can transfer appearance from unconstrained visual exemplars \cite{chen2024referenceimitation}. Recent open-source image editors provide a stronger comparison for subject-preserving generation. LongCat-Image \cite{ma2025longcatimage} and the base JoyAI-Image-Edit interface \cite{song2026joyai} document a single source image, whereas JoyAI-Image-Edit-Plus, FireRed-Image-Edit \cite{qiao2026firered}, and FLUX.2 [klein] \cite{blackforestlabs2025flux2} extend editing to multiple references. Qwen-Image \cite{wu2025qwenimage} supports precise editing and Chinese text rendering but does not document multi-reference conditioning in its technical report. These general editors nevertheless lack the upstream conversion of heterogeneous SKU metadata into verified selling points and a layout policy trained against the downstream generator. Their remaining domain-specific gaps are also visible on CommerceGen-Bench: FLUX.2 [klein] 4B/9B frequently fail on Chinese selling-point text, FireRed loses logos or fuses multiple product conditions unreliably, and Qwen-Image-Edit shows occasional product-identity drift. We state these as observations from our evaluation rather than as limitations claimed by the respective papers. CommerceGen-Studio addresses the full chain jointly: multi-source selling-point extraction, adaptive layout rewriting, dual-reference product-and-trademark conditioning, and character-level text fidelity.

\subsection{Visual Text Rendering}

Diffusion models have long suffered from missing characters, incorrect characters, and character hallucinations in the accurate rendering of complex glyphs. AnyText and AnyText2 \cite{tuo2024anytext,tuo2024anytext2} enhance multilingual text rendering precision through glyph-aware conditioning, but treat text as an independent rendering target while neglecting the coordination among text, overall layout, subject, and trademark, and cannot guarantee that the product subject remains unaltered. TextDiffuser \cite{chen2023textdiffuser} introduces character-level segmentation masks to guide per-character rendering, while its successor TextDiffuser-2 \cite{chen2024textdiffuser2} relaxes this constraint by adopting line-level text regions and LLM-based layout planning for more flexible composition, but still targets single-image text rendering and lack multi-condition joint modeling with product subject and trademark. FLUX-Text \cite{lan2025fluxtext} uses a diffusion Transformer backbone to improve scene text editing but only supports single-image text editing, not multi-condition (white-background image + trademark image + text layout) joint generation. EasyText \cite{lu2026easytext} improves the controllability of multilingual rendering but is not designed for the information consistency needs of e-commerce scenarios and lacks trademark fidelity and product subject constraints. Our work treats text layout as an explicit condition path in multi-condition generation and jointly models it with subject and trademark within a unified diffusion Transformer, filling the above gaps.

\subsection{Reinforcement Learning and Distillation for Diffusion Models}

\begin{figure}[!t]
\centering
\includegraphics[width=\linewidth]{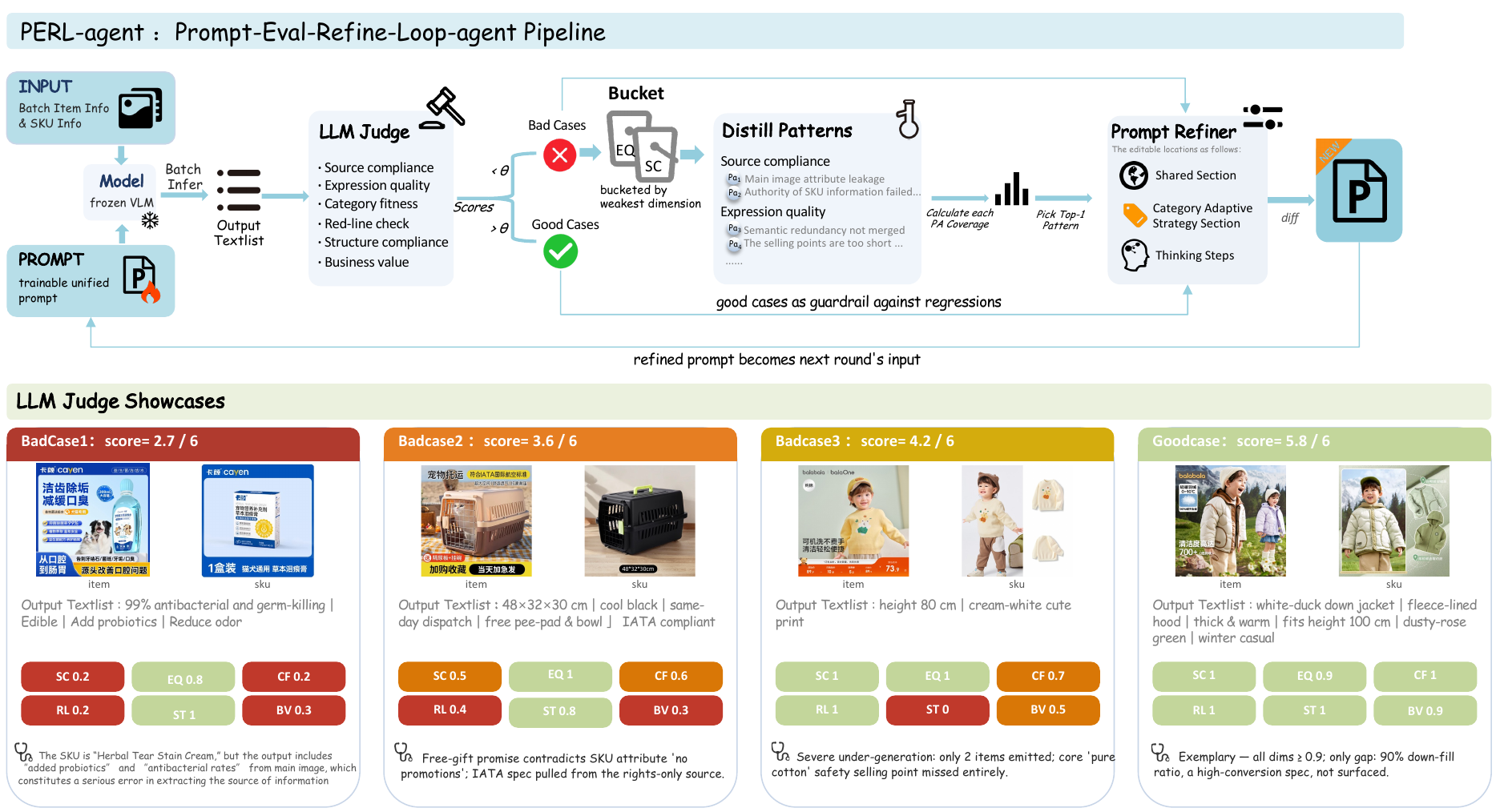}
\caption{Prompt-Eval-Refine-Loop Agent (PERL-Agent): the prompt-evaluation-refinement pipeline. Generator produces selling-point text list  candidates from the current prompt; Judge scores them against a six-dimensional reference-free rubric (SC: Source Compliance $\cdot$ EQ: Expression Quality $\cdot$ CF: Category Fitness $\cdot$ RL: Red-line Check $\cdot$ ST: Structure Compliance $\cdot$ BV: Business Value); Analyzer attributes low-scoring samples by dominant failure dimension to obtain coverage-tagged patterns; Refiner selects the top-1 pattern and applies a minimal single-point diff to produce the next-round prompt.}
\label{fig:perl}
\end{figure}

Reconstruction losses defined at the pixel level say nothing about whether text is legible or a composition is attractive, which is why reinforcement learning has moved into the post-training of diffusion models. Diffusion Transformers and flow-matching objectives \cite{peebles2023scalable,esser2024scaling,lipman2023flowmatching} form the backbone on which current generative systems are built. DDPO \cite{black2024ddpo} applies policy gradients to diffusion post-training by treating denoising as a Markov decision process, where trajectories are sampled on-policy from the current policy and rewards are assigned at the final step. However, it requires per-step policy probability computation and is designed for stochastic diffusion sampling, making it not directly applicable to the deterministic ODE rollout of flow matching. GRPO \cite{shao2024deepseekmath} replaces the learned value function of PPO \cite{schulman2017ppo} with in-group relative advantages; it was designed for language models and does not transfer directly to the continuous-time setting of diffusion and flow matching. Flow-GRPO \cite{liu2025flowgrpo} and DanceGRPO \cite{xue2025dancegrpo} adapt GRPO to flow-matching models and a unified visual generation framework respectively, but the former requires converting a deterministic ODE into a stochastic SDE to inject exploration and introduces additional approximation errors and is prone to reward hacking without KL regularization, and the latter suffers from task interference and seesaw effects in multi-task mixed training. DiffusionNFT \cite{zheng2025diffusionnft} proposes an online reinforcement algorithm compatible with flow matching, performing online reward optimization directly through the diffusion forward process without offline trajectory collection and with stable training; therefore, we adopt it as our policy optimization framework. DiffusionOPD \cite{li2026diffusionopd} provides a diffusion-specific framework for on-policy capability integration: task-specialized teachers are trained independently, while the current student generates its own diffusion trajectories and is distilled toward the corresponding teacher on the states it actually visits. This separates capability-specific exploration from joint integration and reduces interference between heterogeneous objectives.

For inference acceleration, DMD \cite{yin2024dmd} pioneers distribution matching distillation by compressing multi-step sampling to a single step, but single-step distillation yields limited generation quality in complex scenarios. DMD2 \cite{yin2024dmd2} raises few-step quality by refining the distribution matching strategy and adding backward simulation, but does not explicitly protect structural integrity. Based on DMD2, we propose Dynamic Backward Simulation Distillation, which increases the probability of stopping at timesteps near the clean-image endpoint to maintain structural integrity while enhancing generation quality, and introduces discriminator-based adversarial training to strengthen trademark generation, thus balancing quality and latency.

\section{Method}
\subsection{Overview}

CommerceGen-Studio comprises a training pipeline and an inference pipeline, both shown in Figure~\ref{fig:pipeline}. At inference the input is a single product ID. PERL-Agent extracts a structured selling-point text list (TextList) from that SKU's multi-source metadata, and Layout Thinker rewrites the TextList into a layout caption. The text encoder encodes that caption, and SKU-DiT consumes it alongside the white-background product image and the trademark crop to synthesize the final image in one multi-image editing pass. 

Training proceeds in three cascaded stages over our in-house Swift-Image 6B backbone \cite{hu2026swiftimage}. Stage I moves the general backbone onto the e-commerce SKU distribution through pretraining and continual training. Stage II applies Reinforce-Distill-Reinforce to align the generator with human preference and cut inference cost. Stage III trains the VLM layout rewriter under Layout Thinker Optimization so that TextList and SKU-DiT meet through an explicit, executable layout rather than an implicit one.

An item is the parent product listing, and its item image is the listing-level main image; an SKU is a concrete purchasable variant under that item, represented by its own SKU image. For selling-point extraction, we collect and filter heterogeneous information from five sources: text recognized from the item image, text recognized from the SKU image, SKU properties (combining attributes and structured backend fields), the SKU title, and the SKU category. The item image often contains shipping offers and other promotional benefits, but its product attributes may conflict with the selected SKU. We therefore retain useful benefit information while filtering conflicting product facts. For example, if the item image says ``iPhone 17'' but the selected SKU image shows an iPhone 18, preserving ``iPhone 17'' would inject an incorrect selling point into generation. The other four sources are less likely to conflict with the selected SKU, but they are still heterogeneous and verbose; only information that attracts buyers or expresses the product's core value should be retained. PERL-Agent performs this conflict resolution and salient selling-point extraction. Formally, we denote the five-source metadata by $\mathcal{M}=\{T_{item\_ocr},T_{sku\_ocr},P_{properties},T_{title},C_{category}\}$ and map it to a refined selling-point list $L=\pi_{\mathrm{PERL}}(\mathcal{M})$. Through segmentation and detection, we obtain the white-background product image $I_{prod}$ and trademark image $I_{logo}$  respectively; Layout Thinker then rewrites $L$ into $c_{layout}=g_{\phi}(L,I_{prod},I_{logo})$. Our objective is to learn a multi-condition diffusion generation model
\begin{equation}
I_{out} = f_{\theta}\left(I_{prod}, I_{logo}, c_{layout}\right),
\end{equation}
such that $I_{out}$ is a commercial-grade main image directly deployable to production and satisfies: (a) each selling-point fact is supported by the SKU metadata, and the rendered text is character-for-character consistent with the approved TextList; (b) product subject color, texture, quantity, and viewpoint remain unaltered; (c) the trademark does not deform or drift; (d) the layout is typographically coherent at the standard of professionally designed posters.

\subsection{PERL-Agent for TextList Extraction}
\label{sec:perl}

PERL-Agent produces the selling-point text list that drives everything downstream. The conventional alternatives scale poorly: per-category \texttt{CASE WHEN} rules require someone to anticipate every category, and hand-maintained prompts give no way to attribute a regression to a cause or to roll back to a known-good state. We instead treat prompt construction as an optimization problem solved by a closed loop of VLM and LLM roles, scored by a six-dimensional reference-free rubric, with one category-adaptive prompt covering all categories (Figure~\ref{fig:perl}). Appendix~\ref{app:perlbench} specifies the rubric, PERL-Bench-6.

Four roles divide the work: a Generator (VLM) drafts TextList candidates from the current prompt, a Judge scores them against the rubric without reference answers, an Analyzer attributes low scores to a dominant failure dimension, and a Refiner edits the prompt accordingly. We optimize the worst dimension rather than the mean, since a TextList that is fluent but non-compliant is unusable regardless of how well its other dimensions score:
\begin{equation}
\pi^* = \arg\max_\pi \mathbb{E}_{x \sim \mathcal{D}_{\text{cat}}} \left[\min_{d \in \mathcal{R}_6} S_d\left(G_\pi(x), x\right)\right],
\end{equation}
where $G_\pi(x)$ is the TextList generated by Generator using prompt $\pi$ on input $x$, $\mathcal{R}_6$ denotes the 6-dimensional rubric, and $S_d$ is the score on the $d$-th dimension. The prompt is updated by iterative refinement: $\pi^{(t+1)} = \pi^{(t)} \oplus \text{Refine}(\text{Analyze}(\text{Judge}(G_{\pi^{(t)}})))$, where $\oplus$ denotes minimal-diff merging, and Refiner selects the top-1 failure pattern output by Analyzer and applies a single-point modification. Intermediate artifacts are persisted at each round to support comparison and rollback.

\subsection{Stage I: Pretraining \& Continual Training}

The generation backbone adopts our in-house Swift-Image 6B unified multi-image editing Diffusion Transformer \cite{hu2026swiftimage}. We then run a further round of continual training on the SKU vertical domain. Starting from high-quality main images contributed by both premium and long-tail merchants, we recover white-background product images by matting and caption the difference between the white-background and original images, which yields triplets of the form (white-background image, prompt, ground truth). Candidates are filtered twice, by an in-house aesthetics model and by high-saturation color detection, then sampled in a balanced manner across leaf category, background scene type, and in-image text length as described in Section~\ref{sec:trainingdata}. This stage uses approximately 4M product-domain T2I samples, 2.67M single-image I2I editing pairs, and 274K multi-image I2I editing pairs to move the general backbone onto the e-commerce SKU distribution and to stabilize multi-condition text rendering and subject fidelity.

\subsection{Stage II: Progressive Reinforce-Distill-Reinforce Training}
\label{sec:rdr}

Stage I provides SKU-DiT with general multi-condition generation capabilities, but its reconstruction objective does not directly optimize text accuracy, selling-point completeness, subject fidelity, or visual quality. To address these limitations, Reinforce-Distill-Reinforce (RDR) consists of three phases: reinforcement learning aligns the model with human preferences, distillation reduces the number of sampling steps for faster inference, and post-distillation reinforcement recovers the visual details and fidelity lost during distillation. Human preferences rl combines DiffusionNFT-based specialist optimization and DiffusionOPD-based capability integration.

Phase I: Specialist Alignment and Capability Integration. We first define the reward used for DiffusionNFT-based specialist
optimization~\cite{zheng2025diffusionnft}. The reward is evaluated by a
VLM-based judge~\cite{chen2024mllmjudge} along three primary dimensions.
Instruction following ($S_1$) measures text accuracy, selling-point
completeness, trademark restoration, and outpainting compliance.
Content fidelity ($S_2$) evaluates whether the color, texture, quantity,
and viewpoint of the white-background product remain unchanged.
Aesthetic plausibility ($S_3$) assesses shadow grounding, lighting
consistency, clean edges, and typography quality. To further improve
character-level text accuracy, we introduce an additional OCR penalty
computed by PP-OCRv6~\cite{zhang2026ppocrv6}. The resulting reward is

\begin{equation}
\mathcal{R}
=
\sqrt[3]{S_1 \cdot S_2 \cdot S_3}
-
w \cdot \mathrm{penalty}_{\mathrm{OCR}},
\end{equation}

where $w$ controls the OCR penalty strength and
$\mathrm{penalty}_{\mathrm{OCR}}=1$ when an OCR error is detected in
non-product regions and $0$ otherwise. The geometric mean penalizes imbalanced quality across dimensions, so that a serious failure on one dimension substantially reduces the overall reward even when other dimensions score highly.

Using this reward framework, we independently optimize two specialist
teachers with DiffusionNFT. The \emph{complex-layout teacher} is trained
on composition-intensive samples to improve typography, spatial
organization, and multi-element coordination, while the
\emph{subject-preservation teacher} is trained on fidelity-critical
samples to preserve product identity, structure, color, texture, and
trademark details.

After the two teachers converge, we follow the diffusion-specific
on-policy distillation framework of
DiffusionOPD~\cite{li2026diffusionopd} to integrate their complementary
capabilities into a single student SKU-DiT. The student generates its
own diffusion trajectories, and each trajectory is supervised by the
teacher corresponding to its capability domain. This separates
capability-specific exploration from joint capability integration,
allowing the student to acquire both complex-layout generation and
subject preservation without directly mixing the teachers' parameters.
Approximately 5K grouped trajectories are used across specialist
optimization and on-policy capability integration.

Phase II: Distillation. After specialist alignment and on-policy capability integration,the student model has acquired both complex-layout generation and subject-preservation capabilities, but still requires expensive multi-step sampling at every step. Dynamic Backward Simulation Distillation folds both costs into few-step inference. We build on DMD2 \cite{yin2024dmd2} and depart from it in three respects. In the training mixture, general editing data and trademark-domain editing data are combined at a 3:1 ratio; either source alone is worse, as general data underfits domain-specific fonts while trademark-only data yields poorly scaled type. In the sampling schedule, we raise the probability of terminating at timesteps near the clean-image endpoint, which preserves structural integrity that uniform stopping tends to erode, and we add a discriminator trained against high-quality ground-truth images so that trademark strokes are supervised adversarially rather than only in expectation. In the inputs, trademark reference crops are often only tens of pixels across, so we upsample them before conditioning to keep fine stroke detail recoverable. Together with CFG and few-step distillation, single-image latency falls to the 5-second level.

Phase III: Post-Distillation Reinforcement. Distillation buys latency at the cost of a distribution shift, and the damage is not uniform: it concentrates in exactly the high-frequency detail we care most about, blunting font strokes and softening trademark edges. Rather than accept that as the price of speed, we run a second round of reinforcement directly on the few-step student, which re-establishes the subject fidelity and text precision learned during continual training and Phase I, but now under the few-step sampling constraint the student will actually face at inference. This phase uses roughly 2.5K grouped trajectories. The data is drawn from the same pool as continual training with tighter thresholds on aesthetics, composition, and OCR precision -- a smaller and cleaner subset that sits closer to the online distribution than the broad training corpus does.

\subsection{Stage III: Layout Thinker Optimization}

A TextList states what must appear in the image but not where or how, and feeding it to the generator directly forces the generator to invent its own layout, which is what makes composition unstable across samples. Layout Thinker Optimization (LTO) inserts an explicit rewriting layer between the TextList and generator (Figure~\ref{fig:pipeline}). A VLM takes the TextList from PERL-Agent together with the white-background product image and, when available, the trademark crop, and emits a layout caption that decides subject position and orientation, background style and lighting, text hierarchy and color, and trademark position and size. Training proceeds in two stages, supervised fine-tuning (SFT) followed by reinforcement fine-tuning (RFT). Unlike the reinforcement phases of the generator, which are LoRA-adapted \cite{hu2022lora}, both rewriter stages fully fine-tune the upper transformer blocks while keeping the lower blocks frozen. The reason is that layout generation is not a matter of adopting a new output format—a change a low-rank adapter can absorb—but of changing how the model reads a product image: it must re-attend to the subject's extent, saliency, and free space to decide where elements go. In our experiments, low-rank updates were too restrictive to capture this representational shift. The frozen lower blocks preserve the general visual-linguistic grounding from pretraining, and the trainable upper blocks specialize to layout reasoning.

Given the ground-truth layout caption $c^*$, TextList $L$, the white-background product image $I_{prod}$, and the trademark image $I_{logo}$ sampled from the dataset $\mathcal{D}_{SFT}$,
the SFT stage optimizes the VLM by maximizing the likelihood of the ground-truth layout caption, as follows,

\begin{equation}
\mathcal{L}_{SFT} = -\mathbb{E}_{(L, I, c^*)\in \mathcal{D}_{SFT}} \Big[ \frac{1}{T} \sum_{t=1}^{T} \log p_\phi(c_t^* \mid c_{<t}^*, L, I_{prod}, I_{logo}) \Big],
\end{equation}

where $\phi$ denotes the trainable upper-layer parameters of the VLM, $c_t^*$ is the $t$-th token of the caption $c^*$ with length $T$.
To mitigate distribution bias, $\mathcal{D}_{SFT}$ is constructed through balanced sampling across different scenes and product categories. 
At this stage, we primarily emphasize data balancing and prompt engineering, enabling the VLM to acquire the basic mapping capability from structured selling-point information to executable layout descriptions.

The RFT stage aligns the rewriter with both caption quality and rendered-image quality using group-relative policy optimization (GRPO \cite{shao2024deepseekmath}). The text domain scores sampled captions directly on fidelity, information hierarchy, visual richness, and icon feasibility. The image domain renders a separate caption group with the frozen SKU-DiT and scores the resulting images along instruction following, content preservation, visual aesthetics, and logo validity. The two domains normalize rewards within their own rollout groups and define independent KL-regularized policy losses.

For the text-reward domain, the loss is
\begin{equation}
\resizebox{0.98\linewidth}{!}{$\displaystyle \mathcal{L}_{\mathrm{text}}=-\mathbb{E}_{\mathcal{G}_{\mathrm{text}}}\!\left[\frac{1}{N_{\mathrm{text}}}\sum_{i=1}^{N_{\mathrm{text}}}\min\!\left(r_{\mathrm{text},i}(\phi)\hat{A}_{\mathrm{text},i},\operatorname{clip}\!\left(r_{\mathrm{text},i}(\phi),1-\epsilon,1+\epsilon\right)\hat{A}_{\mathrm{text},i}\right)\right]+\beta_{\mathrm{text}}D_{\mathrm{KL}}\!\left(\pi_{\phi}\Vert\pi_{\mathrm{ref}}\right).$}
\label{eq:lto_text_loss}
\end{equation}
For the image-reward domain, the loss is
\begin{equation}
\resizebox{0.98\linewidth}{!}{$\displaystyle \mathcal{L}_{\mathrm{img}}=-\mathbb{E}_{\mathcal{G}_{\mathrm{img}}}\!\left[\frac{1}{N_{\mathrm{img}}}\sum_{i=1}^{N_{\mathrm{img}}}\min\!\left(r_{\mathrm{img},i}(\phi)\hat{A}_{\mathrm{img},i},\operatorname{clip}\!\left(r_{\mathrm{img},i}(\phi),1-\epsilon,1+\epsilon\right)\hat{A}_{\mathrm{img},i}\right)\right]+\beta_{\mathrm{img}}D_{\mathrm{KL}}\!\left(\pi_{\phi}\Vert\pi_{\mathrm{ref}}\right).$}
\label{eq:lto_image_loss}
\end{equation}
At each optimization step, only one reward domain is activated to avoid gradient interference between the text and image objectives. We sample a scalar $u\sim\mathcal{U}[0,1]$ and select the domain according to the configured text-domain probability $p_{\mathrm{text}}$:
\begin{equation}
\mathcal{L}_{\mathrm{RFT}}(u)=
\begin{cases}
\mathcal{L}_{\mathrm{text}}, & u < p_{\mathrm{text}},\\[2pt]
\mathcal{L}_{\mathrm{img}}, & u \ge p_{\mathrm{text}}.
\end{cases}
\label{eq:lto_total_loss}
\end{equation}
Accordingly, each optimization step backpropagates through either the text-domain loss or the image-domain loss, but never both. For $d\in\{\mathrm{text},\mathrm{img}\}$, $\hat{A}_{d,i}=\bigl(R_d(g_{d,i})-\bar{R}_d\bigr)/(\sigma_d+\delta)$ is the reward standardized within domain $d$, while $r_{d,i}(\phi)=\pi_{\phi}(c_{d,i})/\pi_{\phi_{\mathrm{old}}}(c_{d,i})$ is the corresponding importance-sampling ratio. Here, $R_d$ denotes the domain-specific reward, $\delta$ is a numerical-stability constant, $\epsilon$ is the PPO clipping range \cite{schulman2017ppo}, $\beta_d$ controls the domain-specific KL regularization, and $\pi_{\mathrm{ref}}$ is the fixed reference policy. The sampling probability $p_{\mathrm{text}}$ controls the relative optimization frequency of the two domains, with $p_{\mathrm{img}}=1-p_{\mathrm{text}}$. This stochastic domain selection prevents the two reward signals from producing conflicting gradients within the same update.

Throughout RFT, SKU-DiT remains frozen and serves only as the renderer for the image-reward domain. The caption emitted by the VLM is sampled immediately by SKU-DiT, and the resulting image reward updates only the VLM policy. This closes the semantic gap between the rewriter and the fixed generator without interleaving additional SFT steps or modifying SKU-DiT. The RFT-trained VLM is deployed for prompt rewriting at inference.

\section{Dataset and CommerceGen-Bench}

\begin{figure}[!t]
\centering
\includegraphics[width=\linewidth]{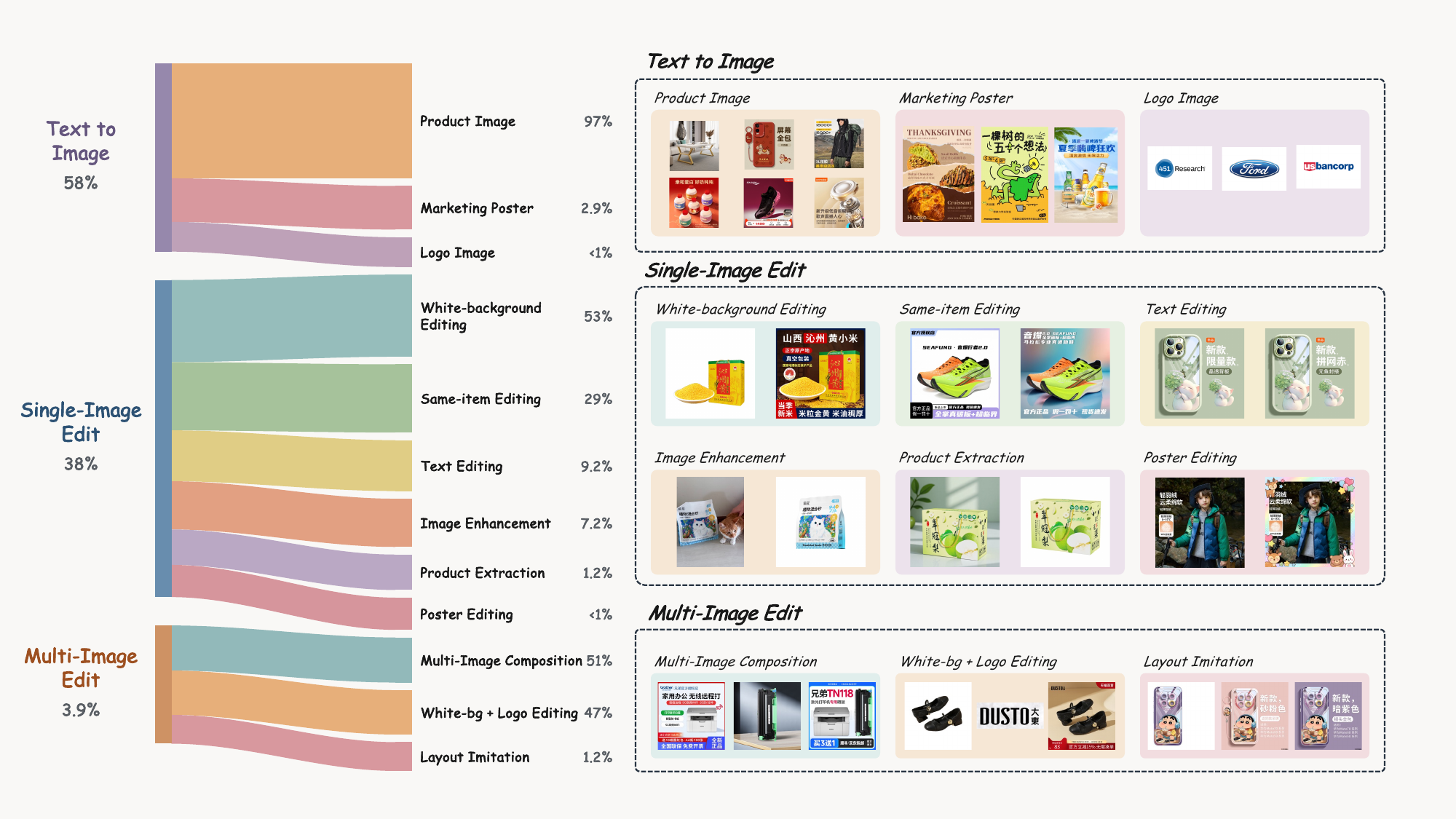}
\caption{\textbf{Task taxonomy of the SKU vertical-domain training corpus.} From left to right: three top-level tasks with their corpus share, their sub-tasks with shares, and representative samples.}
\label{fig:skudata}
\end{figure}

\subsection{Training Model}
\label{sec:trainingmodel}
Our generation backbone is the in-house Swift-Image 6B unified multi-image editing DiT~\cite{hu2026swiftimage}. We initialize it from
the checkpoint obtained after the 500K-step foundational pretraining and 200K-step continual-training schedule reported in Swift-Image~\cite{hu2026swiftimage}. The data curriculum follows the capability-centric design of ~\cite{wang2026corpora}, which progressively increases the training resolution from 256 to 1024 pixels through a five-stage curriculum. We then adapt this checkpoint to the SKU domain using the stage-wise datasets summarized in Table~\ref{tab:data}.

\subsection{Training Data}
\label{sec:trainingdata}

Figure~\ref{fig:skudata} summarizes the task composition of the
SKU-domain training corpus, covering text-to-image generation,
single-image editing, and multi-image editing. The editing data include
both complex-layout samples, which emphasize the organization of
products, selling-point text, trademarks, and decorative elements, and
subject-preservation samples, which require the product identity,
structure, color, texture, and quantity to remain unchanged during
editing.

After task construction, we perform balanced sampling along three
independent dimensions. First, samples are balanced across leaf product
categories to prevent popular categories from dominating the training
distribution. Second, backgrounds are assigned to 50 predefined
background classes, and samples are balanced across these classes to
improve scene diversity. Third, samples are grouped by the number of
selling points in the target image and balanced across groups, ensuring
sufficient coverage of both sparse and text-dense layouts.

For data cleaning, we train an in-house aesthetic scorer based on the
SwinV2-T architecture~\cite{liu2022swinv2} using e-commerce image data.
We combine its aesthetic score with a high-saturation color detector to
remove low-quality or excessively saturated merchant images. This
filtering stage reduces visually undesirable supervision before
balanced sampling and model training.

PERL-Agent uses two data stages. The cold-start stage contains 100 heterogeneous SKU records spanning the five information sources above. It is used to initialize the extraction prompt, expose cross-source conflicts, and establish rules for retaining important selling points. The same 100 records also calibrate the LLM Judge: after the cold-start extractor produces its first-round TextLists, the Judge and multiple domain experts score each result independently. We compute Spearman rank correlation \cite{spearman1904rank} between the sample ordering induced by the mean expert scores and that induced by the Judge scores. Spearman correlation does not require experts to annotate ranks manually; it converts the two sets of scalar scores into ranks and measures whether the Judge orders good and bad TextLists consistently with the experts. A Judge configuration enters the closed loop only when $\rho>0.7$, which serves as a screening threshold for sufficient ranking agreement with domain experts. After cold start and calibration, the scale-up stage expands directly to approximately 1,000 full-category records for prompt optimization and failure-pattern coverage.

\subsection{CommerceGen-Bench}

\begin{figure}[!t]
\centering
\includegraphics[width=\linewidth]{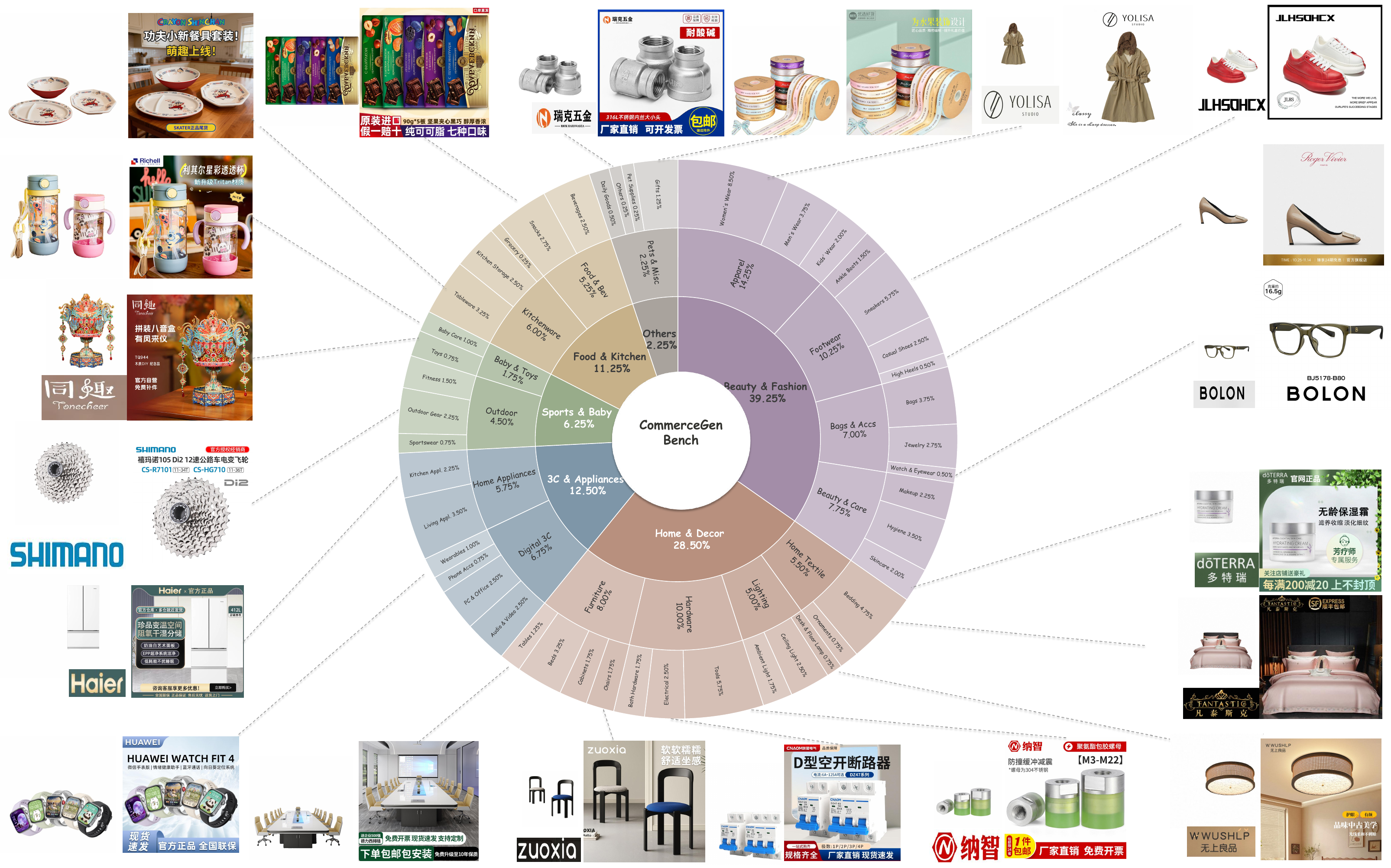}
\caption{\textbf{Distribution of CommerceGen-Bench.} The benchmark comprises 400 common e-commerce scenarios covering 6 top-level categories and 15 sub-categories, with the percentage share of each category visualized.}
\label{fig:benchdist}
\end{figure}

Average-case benchmarks do not adequately reflect model performance in real-world e-commerce scenarios, where difficult and long-tail cases are
particularly important. CommerceGen-Bench comprises 400 real online cases collected from live search result pages, evenly split into 200 single-image editing samples and 200 multi-image editing samples. The benchmark spans 6 top-level categories and 15 subcategories: 3C electronics and home appliances, home goods and home improvement, cosmetics and apparel, general merchandise, sports and maternity, and a residual category. A sample qualifies as hard when at least one of the following holds: the subject has thin or transparent structure that matting tends to destroy, the trademark is small enough that its strokes fall near the resolution limit, or the selling-point text is long enough that character-level errors become likely. Figure~\ref{fig:benchdist} gives the category distribution.

\subsection{Evaluation Protocol}

Recent benchmarks emphasize practical and reasoning-intensive single- and multi-image editing scenarios~\cite{zhou2026cpibench}. Following this evaluation setting, we use Gemini 3 Flash~\cite{google2026gemini3flash} as the multimodal judge model to assess each generated image against its reference product image and editing instruction. To reduce the influence of input resolution, all
images are resized to approximately $1024\times1024$ before evaluation. Five dimensions are assessed on $[0,10]$. Product Fidelity (PF) asks whether the identity, structure, color, and key attributes of the core product match the reference. Instruction Following (IF) asks whether the requested edit was actually carried out. Product Integration (PI) asks whether the edited product sits plausibly in its new scene. Naturalness Detection (ND) measures overall realism and the consistency of lighting and materials. Artifacts Detection (AD) identifies blur, distortion, broken edges, and garbled text. Overall is the arithmetic mean of PF, IF, PI, ND, and AD. Appendix~\ref{app:evalprompt} gives the abridged prompt.

\section{Experiments}

\subsection{Training Configuration}

For matting of white-background product images, we use Mask2Former \cite{cheng2022mask2former} as the segmentation backbone, fine-tuned on e-commerce product data to achieve robust segmentation for complex backgrounds and multi-subject scenarios.

For trademark detection, we use Ultralytics YOLO26 \cite{jocher2026yolo26}, fine-tuned on e-commerce trademark data, to accurately localize and crop trademark regions in main and SKU images. Rendered text is verified throughout with PP-OCRv6 \cite{zhang2026ppocrv6}, which supplies the OCR penalty in the Phase I reward.

\textbf{PERL-Agent.}
PERL-Agent instantiates all four roles with Qwen3.8-27B
\cite{qwen2026qwen38} using role-specific prompts. The Generator
receives both product images and textual SKU metadata and therefore
uses the model's multimodal input capability. The Judge uses the same
backbone in text-only mode: it receives the generated TextList, and the six-dimensional evaluation rubric, but no image input. The Analyzer and Refiner also operate in text-only mode, attributing low scores to failure patterns and updating the extraction prompt, respectively. No model parameters are updated during the
PERL-Agent loop.The refinement loop runs for at most 10 rounds and terminates early when the aggregate rubric score does not improve for three consecutive rounds. In practice, the best score is typically reached within the first three rounds. After termination, we retain the prompt with the highest aggregate rubric score over all completed rounds.

\textbf{Generator.} SKU-DiT is trained as a chain in which each stage initializes from the converged weights of the previous one, and the training method alternates according to what the stage needs.  Since transferring a general backbone onto the product distribution is a change of representation that a low-rank update cannot express, continual training on the SKU vertical domain updates all parameters of the Swift-Image 6B backbone \cite{hu2026swiftimage}. When in RL Phase I, we train two LoRA-based specialist teachers with DiffusionNFT: a complex-layout teacher and a subject-preservation teacher, both using rank 32 and $\alpha=64$. Each teacher uses 8 rollout samples per group, 16 inference steps, and a guidance scale of 4.0. The OCR penalty weight in the reward is set to $w=0.4$ in both reinforcement phases.  After the two specialist teachers converge, their capabilities are integrated into a single student SKU-DiT through DiffusionOPD-based on-policy distillation. This student initializes the full-parameter few-step distillation in Phase II, because folding classifier-free guidance and a multi-step sampler into a few-step student alters the sampling behavior of the whole network rather than adjusting a preference. Phase III post-distillation reinforcement again uses LoRA over the distilled student at the same rank 32 and $\alpha=64$, with 8 rollout samples per group, 8 inference steps to match the student sampler, and a guidance scale of 1.0, guidance having already been distilled into the model by that point.

This alternation between full-parameter updates and low-rank adaptation follows a consistent principle: stages that must acquire or restructure capability update all parameters, whereas the two reward-driven stages are confined to a low-rank subspace. Restricting reinforcement in this way bounds how far reward optimization can pull the model from the grounding established by the preceding supervised or distillation stage, which is the failure mode we most needed to avoid: an unconstrained policy update that chases the aesthetic components of the reward will readily trade away the subject and text fidelity that the earlier stages installed.

\textbf{Layout rewriter.} LTO uses Qwen3.8-27B \cite{qwen2026qwen38} as its backbone. Its reinforcement stage is not LoRA-adapted as the generator's is; instead both LTO stages train the upper transformer blocks in full while the lower blocks stay frozen, so that general visual-linguistic grounding is preserved while the upper blocks specialize to layout planning. Producing a layout requires reasoning jointly over the product image and the TextList to decide placement, hierarchy, and scale, and in our experiments a low-rank update did not give the upper blocks enough capacity to acquire that behavior; the same trainable parameter set is then carried into the RFT stage. In RFT, the two reward domains are activated uniformly at random, i.e., $p_{\mathrm{text}}=0.5$, with each optimization step backpropagating through exactly one domain.

\subsection{Quantitative Comparison}
We conduct a quantitative comparison between CommerceGen-Studio and multiple closed- and open-source models on CommerceGen-Bench; Prompt enhancement(PE) is performed by Qwen3.8-27B \cite{qwen2026qwen38}: in the with-PE setting, all systems share the same enhanced input, whereas in the without 
\begin{table}[!t]
\centering
\setlength{\tabcolsep}{3pt}
\renewcommand{\arraystretch}{0.95}
\resizebox{\linewidth}{!}{%
\begin{tabular}{l c|cccccc|cccccc}
\toprule
& & \multicolumn{6}{c|}{Single-Image Editing} & \multicolumn{6}{c}{Multi-Image Editing} \\
\cmidrule(lr){3-8} \cmidrule(lr){9-14}
Model & Params. & PF & IF & PI & ND & AD & Overall & PF & IF & PI & ND & AD & Overall \\
\midrule
\rowcolor{gray!10}
\multicolumn{14}{l}{\textit{Open-source Models}} \\
Qwen-Image-2.1~\cite{qwen2026qwenimage21} & -- & 8.32 & 6.35 & 8.52 & 7.21 & 6.06 & 7.29 & 7.65 & 5.48 & 8.28 & 6.66 & 5.72 & 6.76 \\
FireRed-Image-Edit-1.0~\cite{qiao2026firered} & 20B & 7.49 & 3.72 & 7.84 & 5.56 & 4.35 & 5.79 & 6.63 & 1.97 & 6.81 & 3.70 & 2.62 & 4.35 \\
Qwen-Image-Edit~\cite{wu2025qwenimage} & 20B & 6.15 & 2.80 & 7.09 & 4.86 & 3.63 & 4.91 & 6.05 & 1.85 & 6.57 & 3.96 & 2.92 & 4.27 \\
FLUX.2 [klein] 9B~\cite{blackforestlabs2025flux2} & 9B & 8.60 & 2.02 & 8.63 & 5.24 & 3.30 & 5.56 & 7.99 & 2.38 & 8.63 & 5.22 & 3.39 & 5.52 \\
FLUX.2 [klein] 4B~\cite{blackforestlabs2025flux2} & 4B & 8.24 & 1.15 & 8.51 & 4.75 & 3.09 & 5.15 & 6.60 & 1.75 & 8.37 & 4.63 & 3.16 & 4.90 \\
JoyAI-Image-Edit-Plus~\cite{song2026joyai} & 16B & 6.95 & 3.56 & 7.59 & 5.78 & 4.56 & 5.69 & 5.74 & 2.85 & 7.72 & 5.52 & 4.33 & 5.23 \\
\midrule
\rowcolor{gray!10}
\multicolumn{14}{l}{\textit{Closed-source Models}} \\
GPT-Image-2~\cite{openai2026gptimage2} & -- & 9.36 & \textbf{9.79} & \underline{9.60} & \underline{9.65} & 9.67 & \textbf{9.61} & \underline{9.39} & \textbf{9.72} & \underline{9.70} & \underline{9.73} & \underline{9.87} & \textbf{9.68} \\
GPT-Image-2.5~\cite{openai2026gptimage25} & -- & \underline{9.36} & \underline{9.63} & \textbf{9.67} & \textbf{9.68} & \underline{9.71} & \underline{9.61} & 9.32 & \underline{9.59} & \textbf{9.73} & \textbf{9.74} & \textbf{9.89} & \underline{9.66} \\
Seedream 5.0 Pro~\cite{bytedance2026seedream5pro} & -- & \textbf{9.44} & 9.47 & 9.50 & 9.61 & \textbf{9.82} & 9.57 & \textbf{9.41} & 9.49 & 9.70 & 9.66 & 9.72 & 9.60 \\
Nano Banana Pro~\cite{google2025nanobananapro} & -- & 8.70 & 9.28 & 9.33 & 9.35 & 9.34 & 9.20 & 8.61 & 8.93 & 9.35 & 9.37 & 9.58 & 9.17 \\
Qwen-Image-2.0-Pro~\cite{alibaba2026qwenimage2pro} & -- & 8.77 & 8.72 & 9.21 & 9.06 & 8.67 & 8.89 & 8.65 & 8.08 & 9.41 & 9.17 & 9.14 & 8.89 \\
Seedream 4.5~\cite{bytedance2026seedream45} & -- & 8.70 & 7.87 & 9.33 & 8.88 & 8.76 & 8.71 & 8.35 & 7.45 & 9.19 & 8.73 & 8.90 & 8.53 \\
\midrule
\color{morandigreen}\textbf{Ours} & \color{morandigreen}6B & \color{morandigreen}9.22 & \color{morandigreen}9.54 & \color{morandigreen}9.42 & \color{morandigreen}9.44 & \color{morandigreen}9.37 & \color{morandigreen}9.40 & \color{morandigreen}9.27 & \color{morandigreen}9.29 & \color{morandigreen}9.54 & \color{morandigreen}9.54 & \color{morandigreen}9.57 & \color{morandigreen}9.44 \\
\bottomrule
\end{tabular}}
\caption{Quantitative results on CommerceGen-Bench without prompt enhancement (PE). \textbf{Bold} and \underline{underline} denote the best and second-best results across all open-source and closed-source baselines; Ours is reported separately and excluded from this ranking. Overall is the mean of PF, IF, PI, ND, and AD within each editing track. Rankings are determined from unrounded values; all displayed values are rounded to two decimal places.}
\label{tab:quantitative}
\end{table}

\begin{table}[!t]
\centering
\setlength{\tabcolsep}{3pt}
\renewcommand{\arraystretch}{0.95}
\resizebox{\linewidth}{!}{%
\begin{tabular}{l c|cccccc|cccccc}
\toprule
& & \multicolumn{6}{c|}{Single-Image Editing} & \multicolumn{6}{c}{Multi-Image Editing} \\
\cmidrule(lr){3-8} \cmidrule(lr){9-14}
Model & Params. & PF & IF & PI & ND & AD & Overall & PF & IF & PI & ND & AD & Overall \\
\midrule
\rowcolor{gray!10}
\multicolumn{14}{l}{\textit{Open-source Models}} \\
Qwen-Image-2.1~\cite{qwen2026qwenimage21} & -- & 8.80 & 9.02 & 9.22 & 9.16 & 8.56 & 8.95 & 8.74 & 8.71 & 9.28 & 9.17 & 8.60 & 8.90 \\
FireRed-Image-Edit-1.0~\cite{qiao2026firered} & 20B & 8.99 & 7.95 & 9.21 & 9.25 & 8.81 & 8.84 & 8.66 & 6.27 & 9.14 & 8.67 & 7.40 & 8.03 \\
Qwen-Image-Edit~\cite{wu2025qwenimage} & 20B & 8.84 & 7.85 & 9.05 & 9.04 & 8.56 & 8.67 & 8.34 & 5.85 & 8.60 & 7.91 & 6.69 & 7.48 \\
FLUX.2 [klein] 9B~\cite{blackforestlabs2025flux2} & 9B & 8.45 & 3.65 & 8.70 & 6.87 & 4.08 & 6.35 & 8.70 & 4.12 & 8.90 & 7.04 & 4.22 & 6.60 \\
FLUX.2 [klein] 4B~\cite{blackforestlabs2025flux2} & 4B & 8.33 & 3.47 & 8.75 & 6.50 & 3.97 & 6.20 & 8.05 & 3.78 & 8.83 & 6.59 & 3.96 & 6.24 \\
JoyAI-Image-Edit-Plus~\cite{song2026joyai} & 16B & 8.65 & 7.90 & 9.18 & 9.09 & 8.63 & 8.69 & 8.40 & 7.20 & 9.31 & 9.17 & 8.50 & 8.52 \\
\midrule
\rowcolor{gray!10}
\multicolumn{14}{l}{\textit{Closed-source Models}} \\
GPT-Image-2~\cite{openai2026gptimage2} & -- & \underline{9.64} & \underline{9.34} & 9.69 & 9.77 & 9.85 & 9.66 & \textbf{9.65} & \textbf{9.30} & \underline{9.74} & \underline{9.79} & \underline{9.92} & \textbf{9.68} \\
GPT-Image-2.5~\cite{openai2026gptimage25} & -- & 9.59 & \textbf{9.52} & \textbf{9.76} & \textbf{9.80} & \textbf{9.92} & \textbf{9.72} & 9.51 & \underline{9.27} & \textbf{9.79} & \textbf{9.81} & \textbf{9.94} & \underline{9.66} \\
Seedream 5.0 Pro~\cite{bytedance2026seedream5pro} & -- & \textbf{9.82} & 9.20 & \underline{9.76} & \underline{9.79} & \underline{9.88} & \underline{9.69} & \underline{9.56} & 9.17 & 9.71 & 9.72 & 9.83 & 9.60 \\
Nano Banana Pro~\cite{google2025nanobananapro} & -- & 9.41 & 8.69 & 9.51 & 9.57 & 9.75 & 9.39 & 9.18 & 8.64 & 9.50 & 9.54 & 9.79 & 9.33 \\
Qwen-Image-2-Pro~\cite{alibaba2026qwenimage2pro} & -- & 8.95 & 8.86 & 9.21 & 9.35 & 9.16 & 9.11 & 9.19 & 8.50 & 9.61 & 9.64 & 9.78 & 9.34 \\
Seedream 4.5~\cite{bytedance2026seedream45} & -- & 9.20 & 8.52 & 9.45 & 9.47 & 9.54 & 9.24 & 9.03 & 8.18 & 9.48 & 9.44 & 9.63 & 9.15 \\
\midrule
\color{morandigreen}\textbf{Ours} & \color{morandigreen}6B & \color{morandigreen}9.48 & \color{morandigreen}8.83 & \color{morandigreen}9.63 & \color{morandigreen}9.66 & \color{morandigreen}9.55 & \color{morandigreen}9.43 & \color{morandigreen}9.60 & \color{morandigreen}8.72 & \color{morandigreen}9.58 & \color{morandigreen}9.62 & \color{morandigreen}9.70 & \color{morandigreen}9.45 \\
\bottomrule
\end{tabular}}
\caption{Quantitative results on CommerceGen-Bench with prompt enhancement (PE). The notation follows Table~\ref{tab:quantitative}; Ours is reported in green and excluded from baseline ranking.}
\label{tab:quantitative_pe}
\end{table}

\begin{table}[!t]
\centering
\begin{minipage}[t]{0.47\linewidth}
\centering
\vspace{0pt}
\footnotesize
\setlength{\tabcolsep}{2pt}
\begin{tabular*}{\linewidth}{@{\extracolsep{\fill}}llc@{}}
\toprule
Layout Model & Generation Model & Score \\
\midrule
VLM base & SKU-DiT (50 steps) & 6.509 \\
VLM base + SFT & SKU-DiT (50 steps) &  7.832 \\
VLM base + SFT & RD SKU-DiT (50 steps) &  8.296 \\
VLM base + SFT & RD SKU-DiT (8 steps) & 8.253 \\
VLM base + SFT & RDR SKU-DiT (8 steps) & 8.271 \\
\textbf{LTO VLM} & \textbf{RDR SKU-DiT (8 steps)} & \textbf{9.440} \\
\bottomrule
\end{tabular*}
\caption{Ablation study on CommerceGen-Bench. RDR means Reinforce-Distill-Reinforce, RD means Reinforce-Distill.}
\label{tab:ablation}
\end{minipage}%
\quad
\begin{minipage}[t]{0.49\linewidth}
\centering
\vspace{0pt}
\footnotesize
\setlength{\tabcolsep}{2pt}
\begin{tabular*}{\linewidth}{@{\extracolsep{\fill}}lll@{}}
\toprule
Stage & Data Type & Scale \\
\midrule
CT: Product T2I & T2I triplets & $\sim$4.07\,M \\
CT: Single-image I2I & Editing pairs & $\sim$2.67\,M \\
CT: Multi-image I2I & Editing pairs & $\sim$274\,K \\
LTO: SFT  & Layout captions & $\sim$20\,K \\
LTO: RFT & Prompt groups & $\sim$5\,K \\
RDR I: RL & Group traj. & $\sim$5\,K \\
RDR II: Distill. & Mixed data & $\sim$200\,K \\
RDR III: RL & Group traj. & $\sim$2.5\,K \\
PERL-Agent & Cold start / scale-up & 100 / 1K \\
\bottomrule
\end{tabular*}
\caption{Data scale of each training stage.}
\label{tab:data}
\end{minipage}
\end{table}

\begin{figure}[H]
\centering
\includegraphics[width=0.99\linewidth,keepaspectratio]{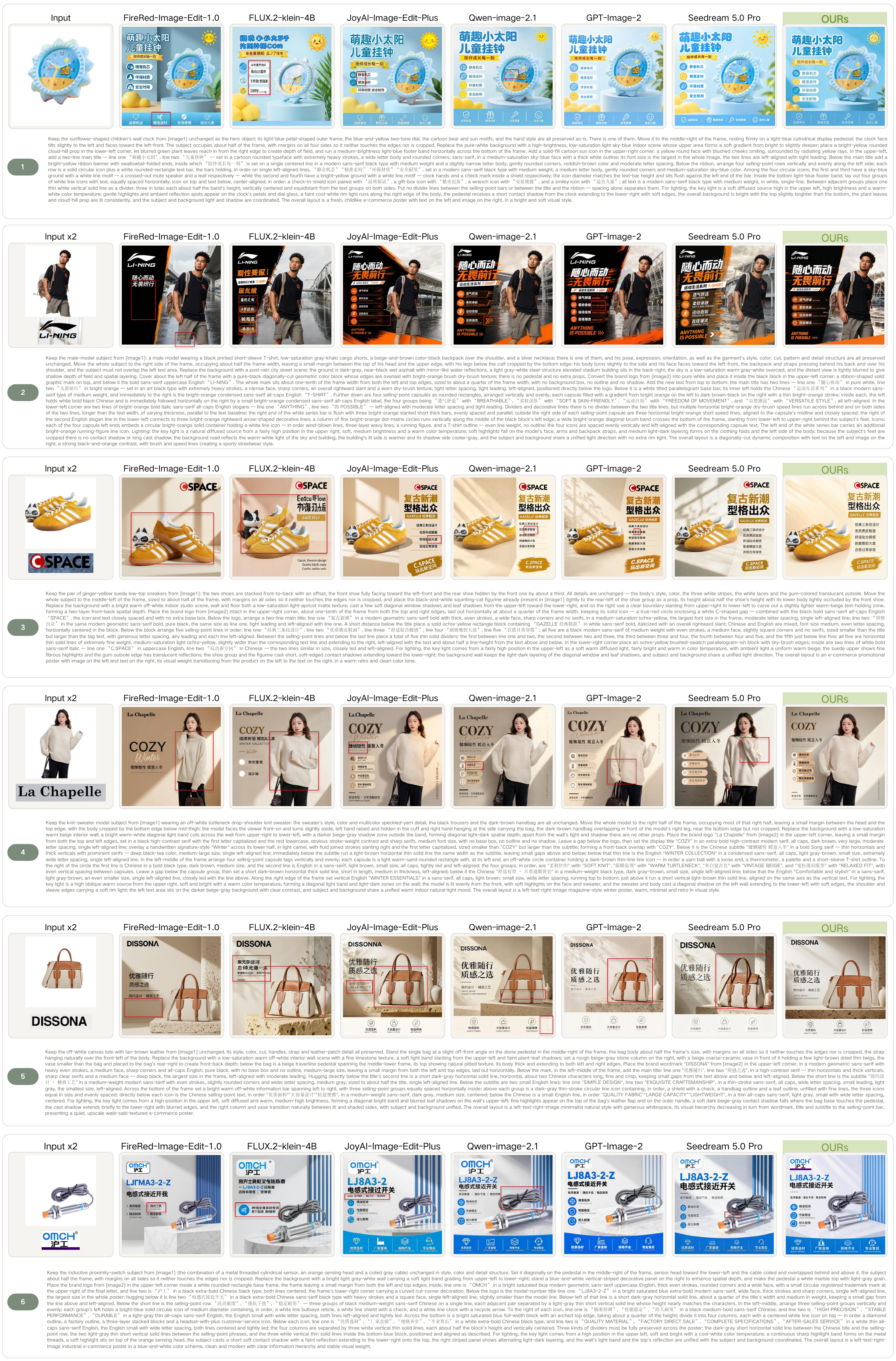}
\caption{Qualitative comparison of CommerceGen-Studio and several models on single-image editing and multi-image editing.}
\label{fig:qualitative}
\end{figure}

\clearpage
\begin{figure}[H]
\centering
\includegraphics[width=0.99\linewidth,keepaspectratio]{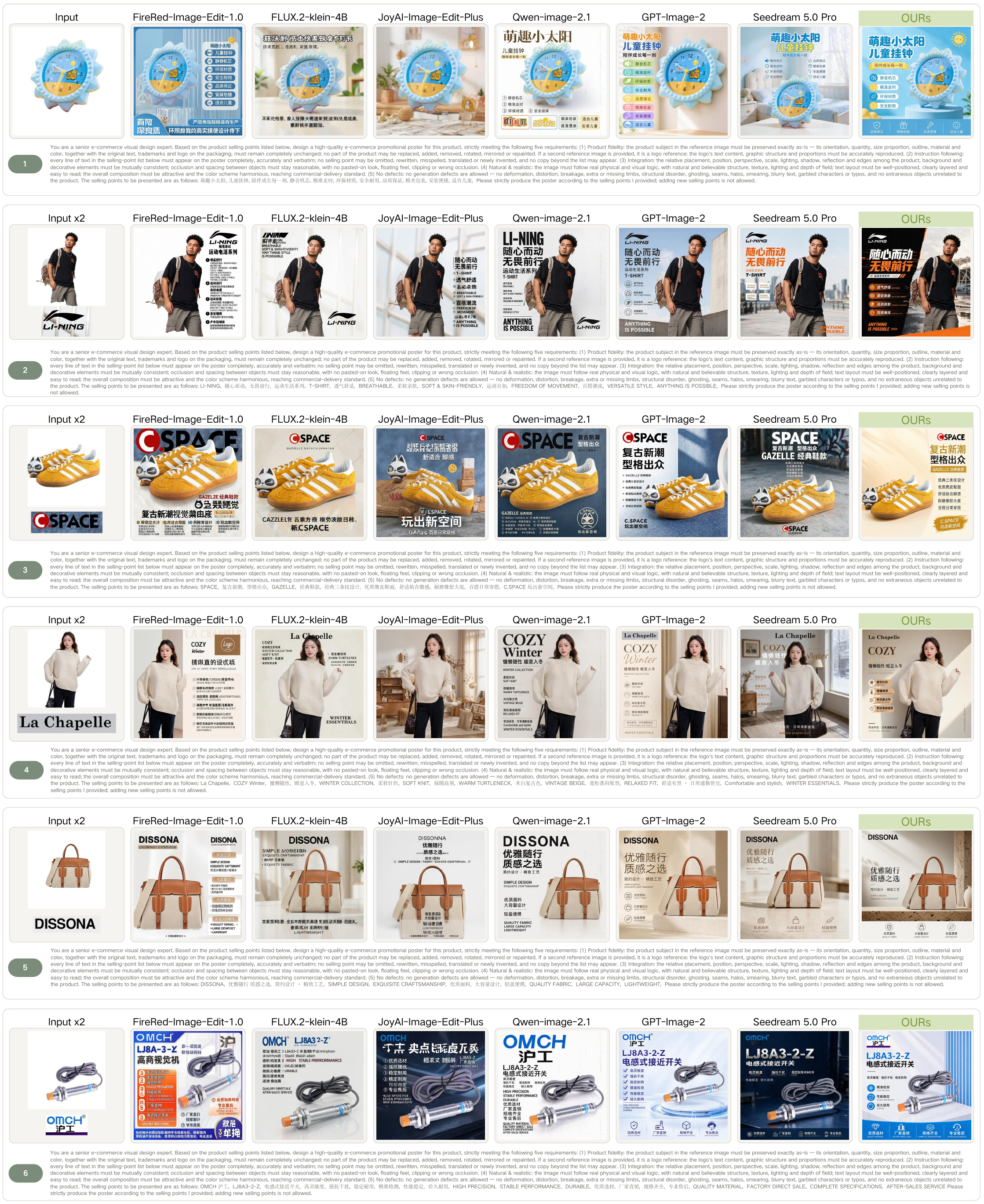}
\caption{Qualitative results of several models without prompt enhancement (PE).}
\label{fig:qualitative_nope}
\end{figure}

Results without and with PE are shown in Tables~\ref{tab:quantitative} and~\ref{tab:quantitative_pe}, respectively. the exact without-PE prompt format is shown in Figure~\ref{fig:qualitative_nope}. 

-PE setting we use the original instruction that directly contains the selling-point TextList and basic requirements; Without PE, CommerceGen-Studio achieves Overall scores of 9.40 and 9.44 on the single and multi-image tracks, respectively; with PE, the scores are 9.43 and 9.45. Among open-source baselines, Qwen-Image-2.1 is strongest overall in both settings, reaching 7.29/6.76 without PE and 8.95/8.90 with PE on the single-/multi-image tracks, but it remains below CommerceGen-Studio in product fidelity, integration, and artifact control. GPT-Image-2 and GPT-Image-2.5 provide the strongest closed-source results. Comparing Tables~\ref{tab:quantitative} and~\ref{tab:quantitative_pe} shows that PE substantially improves instruction following for most baselines, while domain-specific training remains critical for faithful product-and-trademark composition.

For Ours, the enhanced prompt introduces additional explicit placement, hierarchy, and style requirements, thereby increasing the number of verifiable constraints. Consequently, cross-setting IF changes reflect both generation quality and prompt difficulty and should not be interpreted as a pure capability change.

\subsection{Qualitative Examples}

Figures~\ref{fig:qualitative} and~\ref{fig:qualitative_nope} complement the quantitative results with representative single and multi-image editing cases. Figure~\ref{fig:qualitative} shows that CommerceGen-Studio preserves product appearance, logos, and selling-point text more consistently than general editors, particularly when product and trademark references must be fused. Figure~\ref{fig:qualitative_nope} isolates the compared models without PE: the most common failures are product-identity drift, omitted or corrupted Chinese text, and unstable multi-reference composition. These visual patterns are consistent with Tables~\ref{tab:quantitative} and~\ref{tab:quantitative_pe}, where PE improves instruction following for several baselines but does not eliminate domain-specific fidelity and integration errors.

\FloatBarrier

\subsection{Ablation Study}

To assess the contribution of each training stage, we start from an open-source VLM baseline and the SKU-DiT base model, and progressively add each training stage; the results are shown in Table~\ref{tab:ablation}. The data scale of each SKU vertical-domain training stage is summarized in Table~\ref{tab:data}. The score is computed on 400 samples from CommerceGen-Bench with PE.

The measured endpoints show a substantial gap between the open-source layout baseline(6.509) and the complete CommerceGen-Studio configuration (9.440). SFT provides the first substantial gain by adapting the layout
rewriter to the e-commerce domain. Distillation enables 8-step sampling with a small score decrease from 8.296 to 8.253, after which post-distillation reinforcement partially recovers the loss to 8.271. LTO then provides the largest late-stage gain, raising the score from 8.271 to 9.440.

\section{Conclusion and Future Work}

We presented CommerceGen-Studio, an end-to-end framework for producing commercial-grade main images at SKU granularity. Its three components address the four requirements we set out with: PERL-Agent extracts selling points from heterogeneous metadata under a reference-free rubric, Layout Thinker Optimization turns those selling points into layouts the generator can execute, and SKU-DiT trained under Reinforce-Distill-Reinforce renders them with a latency of a few seconds without surrendering fidelity. On CommerceGen-Bench, our hard-sample benchmark, the framework achieves the highest Overall scores among the evaluated open-source baselines on both editing tracks, with and without PE, and remains competitive with leading closed-source systems, outperforming several of them on selected dimensions. Two limitations point to future work. First, our reward signals are defined by rubrics and therefore capture what we can articulate; incorporating real user behavior would let the model learn aesthetic preferences that resist explicit specification. Second, the system currently optimizes for a single notion of quality per SKU, and conditioning on audience segment and style would let one product yield genuinely different assets rather than one canonical treatment.

\appendix

\section{Appendix}
\label{app:main}

\subsection{Evaluation Prompt}
\label{app:evalprompt}

Figure~\ref{fig:evalprompt} reproduces the rubric given to the judge model, abridged here for space. Each dimension additionally carries per-band anchor descriptions at the 10, 9, 7--8, 5--6, 3--4, 1--2, and 0 levels, written in the same conservative style as the calibration rules below; we omit those anchor tables and retain the instructions that govern scoring behavior.

\begin{figure}[H]
\centering
\fbox{\begin{minipage}{0.95\linewidth}
\scriptsize
\textbf{Evaluation Prompt (Forensic Rubric Scoring)}\\[3pt]
\texttt{You are a conservative, forensic evaluator of commercial product image editing. Compare every visible detail in the generated image against the reference image(s) and the editing instruction. Return ONLY one JSON object.}\\[6pt]
\textbf{CALIBRATION RULES}
\begin{itemize}
\setlength{\itemsep}{1pt}\setlength{\parskip}{0pt}
\item Score five dimensions independently from 0 to 10; decimals are allowed and higher is better.
\item Start each dimension at 5.0, then move it only when visible evidence satisfies an anchor. Do not start from 10 and subtract.
\item 5 means mixed/borderline, not catastrophic. 7 means good but clearly flawed. 8 means strong with limited defects. 9 means excellent with only negligible defects. 10 means essentially flawless after close inspection and must be rare.
\item Never award 9 or 10 from a favorable overall impression alone. A score above 8 requires naming why there is no material defect in that dimension.
\item When a relevant small detail or text is visibly unreadable, do not assume it is correct. Do not reward beauty, resolution, or style in dimensions where it does not prove correctness.
\item Use the lower end of a band when several defects exist and the upper end when only one limited defect exists. When uncertain between two scores, choose the lower score.
\end{itemize}
\textbf{DIMENSIONS}
\begin{itemize}
\setlength{\itemsep}{1pt}\setlength{\parskip}{0pt}
\item \textbf{PF -- Product Fidelity:} identity, silhouette, geometry, proportions, materials, colors, fine details, original packaging text, trademarks, and product logos. Layout or camera changes alone are not fidelity errors.
\item \textbf{IF -- Instruction Following:} completion and correctness of every explicit requested edit, object, placement, count, relation, style, and newly added or modified text.
\item \textbf{PI -- Product Integration:} how coherently the preserved product and optional logo belong in the edited composition, including placement, scale, perspective, lighting, shadows, reflections, edges, occlusion, and scene interaction.
\item \textbf{ND -- Naturalness Detection:} perceptual realism and visual plausibility appropriate to the requested photo, advertising, illustration, or graphic-design style. Judge anatomy, physics, texture, lighting, depth, typography placement, and composition, not mere attractiveness.
\item \textbf{AD -- Artifacts Detection:} absence of malformed geometry, duplicated or missing parts, garbled pixels or text, halos, seams, smearing, ghosting, unintended objects, broken edges, and generation residue. A higher score means fewer artifacts.
\end{itemize}
\rule{\linewidth}{0.4pt}\\[2pt]
\textbf{Overall Score:} the arithmetic mean of the five dimension scores, $(\text{PF}+\text{IF}+\text{PI}+\text{ND}+\text{AD})/5$.
\end{minipage}}
\caption{Forensic evaluation prompt for CommerceGen-Bench (abridged): a conservative five-dimensional rubric (PF, IF, PI, ND, AD) scored 0--10, where the overall score is the arithmetic mean of the five dimensions. Per-band anchor descriptions are omitted for space.}
\label{fig:evalprompt}
\end{figure}

\subsection{PERL-Bench-6}
\label{app:perlbench}

The six-dimensional ground-truth-free rubric used to evaluate PERL-Agent's TextList quality is defined as follows: Source Compliance (SC) measures the consistency between text information and SKU metadata; Expression Quality (EQ) measures the fluency and professionalism of language expression; Category Fitness (CF) measures how well the content matches category-specific characteristics; Red-line Check (RL) detects whether the content violates rules or contains sensitive material; Structure Compliance (ST) measures the structural regularity of the text list; Business Value (BV) measures the commercial conversion potential of selling-point information. Each dimension is scored from 0 to 1, and a rule module applies reverse penalty scoring for specific patterns. By default, an aggregate score $\le 5$ out of 6 is classified as a bad case, triggering Analyzer attribution and Refiner rollback.

\clearpage
\bibliography{references}
\bibliographystyle{colm2024_conference}

\end{document}